\documentclass[conference]{IEEEtran}

\IEEEoverridecommandlockouts                              % This command is only needed if 
\usepackage{siunitx}
\usepackage{layouts}
\usepackage{scalerel}
\usepackage{graphicx}
\usepackage{standalone}
\usepackage{pgfplots}

\let\labelindent\relax
\usepackage{enumitem}
\usepgfplotslibrary{groupplots, statistics}
\usepackage{tikz}
\usetikzlibrary{shapes,backgrounds,shapes.geometric}
\usepackage{circuitikz}
\tikzset{font={\fontsize{9pt}{9}\selectfont}}
\usepackage{wrapfig}
\usepackage{multirow}
\usepackage{makecell}
\usepackage{amsmath,amssymb,amsfonts}
\usepackage{bm}
\usepackage{subfigure}

\usepackage{xstring}

\usepackage{jabbrv}

\usepackage{tabularx}
\usepackage{makecell}
\usepackage{multirow}
\usepackage{booktabs}
\newcolumntype{Z}{>{\centering\arraybackslash}X}
\makeatletter
\let\NAT@parse\undefined
\makeatother
\usepackage{hyperref}  %hyperref still needs to be put at the end!

\title{\LARGE \bf Studying the Effect of Explicit Interaction Representations on
Learning Scene-level Distributions of Human Trajectories
}

\author{Anna M\'esz\'aros$^*$, Javier Alonso-Mora, and Jens Kober% <-this % stops a space
\thanks{*Corresponding author}% <-this % stops a space
\thanks{This research was supported by NWO-NWA project “Acting under uncertainty” (ACT), NWA.1292.19.298.}
\thanks{All authors are with the Cognitive Robotics Department,
        TU Delft, 2628 CB Delft, The Netherlands
        {\tt\small \{A.Meszaros, J.AlonsoMora, J.Kober\}@tudelft.nl}}%
}

\begin{document}

\maketitle
\thispagestyle{empty}
\pagestyle{empty}

%%%%%%%%%%%%%%%%%%%%%%%%%%%%%%%%%%%%%%%%%%%%%%%%%%%%%%%%%%%%%%%%%%%%%%%%%%%%%%%%
\begin{abstract}
Effectively capturing the joint distribution of all agents in a scene is relevant for predicting the true evolution of the scene and in turn providing more accurate information to the decision processes of autonomous vehicles.
While new models have been developed for this purpose in recent years, it remains unclear how to best represent the joint distributions particularly from the perspective of the interactions between agents.
Thus far there is no clear consensus on how best to represent interactions between agents; whether they should be learned implicitly from data by neural networks, or explicitly modeled using the spatial and temporal relations that are more grounded in human decision-making. %there is no clear consensus on whether interactions and in turn the joint distributions should be implicitly captured from data by neural networks, or whether they should be captured explicitly on the basis of the agents' spatial and temporal relations to each other.
This paper aims to study various means of describing interactions within the same network structure and their effect on the final learned joint distributions.
Our findings show that more often than not, simply allowing a network to establish interactive connections between agents based on data has a detrimental effect on performance. 
Instead, having well defined interactions (such as which agent of an agent pair passes first at an intersection) can often bring about a clear boost in performance.

\end{abstract}

%%%%%%%%%%%%%%%%%%%%%%%%%%%%%%%%%%%%%%%%%%%%%%%%%%%%%%%%%%%%%%%%%%%%%%%%%%%%%%%%
\section{INTRODUCTION}
Autonomous vehicles (AVs) promise to bring about safer and more accessible roads~\cite{brar2017impact, meyer2017autonomous}. 
However, for an AV to navigate in an environment with people, from drivers to vulnerable road users such as cyclists and pedestrians, AVs require the capacity to anticipate what the people around them will do.
Predicting human behavior in traffic, however, is complicated by the fact that such behavior is generally not deterministic, but stochastic, taking the form of potentially complex and multi-modal distributions~\cite{ferro2020stochastic}. 

A possible way to approach predicting the future trajectories of agents is by predicting each individual agent in the scene based on context information such as the past trajectories of agents in the scene and static environment information~\cite{salzmann2020trajectron++}. %, deo2022multimodal
These types of prediction models provide what are formally known as marginal predictions.
However, a drawback of these types of models is that they fail to account for interactions between agents in the future, resulting in misinterpretations of the actual likelihood of certain outcomes -- e.g. two vehicles colliding -- when observing the evolution of a scene as a whole~\cite{luo2023jfp}.
Recently, new prediction models have been developed to address the matter of joint predictions of future trajectories of all agents in a scene. 
A number of these models leverage multi-head attention or transformer networks to capture the interaction between agents and jointly predict their future trajectories~\cite{aydemir2023adapt, girgis2021latent}.
Meanwhile, other models utilize learned interaction graphs to capture interactions between agents and factorize the joint distribution, in turn simplifying the learning process such as in the FJMP model~\cite{rowe2023fjmp}.
% Nevertheless, a common feature of all these models is that they predict a fixed number of trajectories -- generally associated to individual scene-level modes -- which limits their usability for risk-aware motion planners~\cite{mustafa2024racp}.

The way in which interactions between agents should be provided in machine learning approaches remains under-explored, despite being a key factor in human trajectory prediction~\cite{benrachou2022use}.
Often, these interactions are captured implicitly by the network itself~\cite{aydemir2023adapt, 
girgis2021latent}.
% shi2024mtr++
% jiang2023motiondiffuser, 
% zhou2022hivt
% kamenev2022predictionnet, 
% wang2024joint, 
% pu2024lightweight, 
% bi2019joint, 
% schafer2024caspnet++, 
% cho2019deep, 
% jia2022multi, 
% xin2025multi, 
% wang2024optimizing, 
% zhu2021star, 
% zhao2019multi, 
 In other cases, agents are connected by fully connected graphs whose edges carry agents' relative distance, velocity, or direction~\cite{li2024efin, chen2023goal, mo2022multi}. 
 % ma2021continual, 
 % cui2021lookout
 % liu2022multi
 Other works capture interactions by constructing agent cliques~\cite{chen2022scept, nivash2024simmf}. 
 However, these approaches only jointly predict agents within cliques and ignore inter-clique interactions. 
 Existing extensions remain limited to homogeneous groups (typically pedestrians)~\cite{chen2022hgcn}. 
 Meanwhile other approaches leverage spatial heuristics such as the crossing of future trajectories~\cite{rowe2023fjmp, sun2022m2i}, hypothetical conflicts~\cite{bhattacharyya2024ssl, zhang2024edge}, collision risk~\cite{xue2025rethink}, %~\cite{cui2024dbn, chai2025gacnet, xue2025rethink, he2024vehicle}, 
 movement direction (i.e whether agents are moving towards or away from each other)~\cite{wen2024second}, Euclidean distance~\cite{
 kang2024ffinet, 
 yuan2021agentformer}, 
 % woo2024fimp, 
 % tang2023fgnet, 
 %chen2025adaptive,
 %hu2025hypergraph,
 % yao2024attention, 
 % wei2025goal, 
 % zhou2025heterogeneous, 
 % guo2023query, 
 % zhang2022trajectory, 
 or social forces~\cite{mo2024pi} to provide more structure for learning these interactions.
 More abstract heuristics such as feature
 similarity~\cite{zhu2023ipcc}
 % , cao2022leveraging} 
 or correlation~\cite{xu2022groupnet} have also been used to guide interaction learning.
While prior work has explored agent interaction modeling~\cite{diehl2019graph}, these approaches are largely limited to marginal prediction models with homogeneous agents.
To our knowledge, no work has yet investigated which interaction representation is the most beneficial for machine learning models for joint human trajectory prediction, particularly for heterogeneous agents.

In this study, we examine how the chosen interaction representation affects not only the accuracy of individual trajectory predictions, but also the model’s ability to capture distributions over multiple plausible future trajectories, which is crucial for risk-aware motion planning~\cite{mustafa2024racp}.
A promising group of models for capturing complex and even multi-modal distributions from data are Normalizing Flows (NFs)~\cite{tabak2013family}. %tabak2010density, 
This family of methods has already been successfully applied for predicting marginal distributions over agent trajectories~\cite{scholler2021flomo, meszaros2024trajflow}.
Nevertheless, expanding these approaches for predicting joint distributions of all agents in a scene is not straightforward due to the bijective transformation functions that make up the core of NFs which require the dimensionality of the input to a NF to be kept constant.
This results in complications since the number of agents in a scene often changes. 
A naive solution to this problem would be to introduce dummy agents and only focus on a fixed number of agents in a scene.
However, this can lead to a number of issues, such as wasted computational resources if too many dummy agents are used to ensure the maximum number of agents is covered.
Alternatively, one could use heuristics to determine the most important agents to predict, which may however result in relevant agents being overlooked, particularly in more crowded scenarios.
In response to these challenges we leverage the idea of factorizing the joint distribution in accordance to a directed acyclic graph (DAG) which captures the direction of interaction between agents. 

The contributions of our work are two-fold. 
Firstly, we propose a Graph-based Motion Prediction (GMoP) model structure based on normalizing flows to capture joint distributions over the trajectories of all agents in a scene which remains flexible to the varying number of agents from scene to scene.
Secondly, we use this model structure to study the manner in which interactions between heterogeneous agents~\footnote{Scenes can include a combination of up to four different agent types -- vehicles, motorcyclists, cyclists and pedestrians.} might be modeled on the level of the DAG and how these affect the final learned distributions.
We perform the study on four popular real-world driving datasets (Argoverse~\cite{wilson2023argoverse}, INTERACTION~\cite{zhan2019interaction}, nuScenes~\cite{caesar2020nuscenes}, and rounD~\cite{rounDdataset}).

\section{Background: Normalizing Flows}
Normalizing Flows are a generative method, capable of capturing complex distributions. 
NFs are based on the concept of transforming distributions through a series of differentiable bijective functions into a simple known ``base'' distribution $Z_0$ -- most commonly a standard normal distribution.
One type of flow model is the conditional auto-regressive NF~\cite{lu2020structured}, %~\cite{papamakarios2021normalizing}, 
consisting of a series of $K$ normalizing layers, and conditioned by a feature vector $C$ that can take the form of, e.g., an encoding of observations like past trajectories, static environment, and social interactions. 
% The main components of these layers are the conditioner~$c_k$ and the transformer~$\tau_k$, which are often accompanied by an additional permutation layer~$\epsilon_k$.
% The latter two functions ($\tau_k$ and $\epsilon_k$) are bijective -- and therefore invertible. 
% In the generative direction, these functions enable the transformation of a sample $\bm{z}_0$ from the base distribution $Z_0$ into the desired distribution $Z_K$:
% \begin{equation*}
%     % z_i^\prime = \tau_i (z_i, h_i), \quad \text{where} \quad h_i=c_i(z_{\leq i}), 
%     \bm{z}_{k+1} = \epsilon_k\left(\tau_k (\bm{z}_k; \bm{\theta}_k)\right), \quad \text{with} \quad \bm{\theta}_k=c_k\left(\bm{z}_{k};C\right), \label{eq:NF_Layer}
% \end{equation*}
% where $\bm{z}_{k+1}$ is the result of the $k$-th intermediate transformation. 
% Meanwhile, $C$ is a conditioning input~\cite{lu2020structured} that can take the form of e.g. an encoding of observations like past trajectories, static environment, and social interactions.
% In the normalizing direction, $F$ is then a composition of all $K$ layers, where it is possible to exploit the property of $c_k$ that $\bm{\theta}_k = c_k\left(\epsilon_k^{-1}(\bm{z_{k+1}});C\right)$:
% \begin{equation*}
%     F(\bm{z}_K) = \left(\tau^{-1}_0 \circ \epsilon^{-1}_0 \cdots \circ \tau^{-1}_K \circ \epsilon^{-1}_K \right) (\bm{z}_K) = \bm{z}_0
% \end{equation*}

In the normalizing direction, the normalizing layers make up the normalizing flow function $F$ which transforms samples $z_K$ from the desired distribution $Z_K$ into the base distribution $Z_0$.
In the generative direction, it is then possible to draw a sample ${\bm{z}_K = F^{-1}(\bm{z}_0)}$ from the desired non-normal distribution over outputs $Z_K$, using a sample $\bm{z}_0$ from $Z_0$. 
The Probability Density Function (PDF) $p_K$ of $Z_K$ can then also be obtained in terms of the PDF $p_0$:
\begin{equation*}
\begin{split}
    p_K(\bm{z}_K) &= p_0(F(\bm{z}_K)) \, |\det J_{F}(\bm{z}_K)| \\
                    &= p_0(\bm{z}_0) \, |\det J_{F^{-1}}(\bm{z}_0)|^{-1},
\end{split}
\end{equation*}
The absolute determinant of the Jacobian $|\det J_F(\bm{z}_K)|$ quantifies the relative change of volume within a small neighborhood of $\bm{z}_K$ when transforming it to a sample $\bm{z}_0$ using $F$.
This ensures that the probability mass remains the same between the two distributions.
The parameters of $F$ are learned by minimizing the negative log-likelihood over a finite number $N$ of training samples $z_{K,n}$:

\begin{equation*}
    \mathcal{L} \approx -N^{-1}{\textstyle \sum^{N}_{n=1}\log p_0(F(\bm{z}_{K,n}))} + \log |\det J_{F}(\bm{z}_{K,n})|.
\end{equation*}

% The parameters of $F$ are learned by minimizing the KL-divergence between the target distribution~$Z_K^{*}$ with PDF $p_K^{*}(\bm{z}_K)$ and the learned distribution $Z_K$ with the PDF $p_K(\bm{z}_K)$:
% % \begin{equation}
% \begin{align*}
%     \mathcal{L} =& D_{\mathrm{KL}}[p_K^{*}(\bm{z}_K)||p_K(\bm{z}_K)]\\
%                 =& -\mathbb{E}_{\bm{z}_K\sim Z_K^*}\left[\log p_0(F(\bm{z}_K)) +\log |\det J_{F}(\bm{z}_K)| \vphantom{\log p_K^{*}(\bm{z}_K)}\right. \\
%                 &\phantom{-\mathbb{E}_{\bm{z}_K\sim Z_K^*}}
%                 \; \left. \vphantom{\log p_0(F(\bm{z}_K)) +\log |\det J_{F}(\bm{z}_K)|}
%                 - \log p_K^{*}(\bm{z}_K)\right]\notag\\ 
% \intertext{With only a finite number $N$ of samples $\bm{z}_{K, n}$ representing the underlying distribution $Z_K^{*}$ and ignoring the constant part $\log p_K^{*}(\bm{z}_K)$, this loss can be approximated with: }
%     \mathcal{L} \approx& -\frac{1}{N}\sum^{N}_{n=1}\log p_0(F(\bm{z}_{K,n})) + \log |\det J_{F}(\bm{z}_{K,n})|.
% \end{align*}

\begin{figure*}
    \centering
    \resizebox{\textwidth}{!}{\input{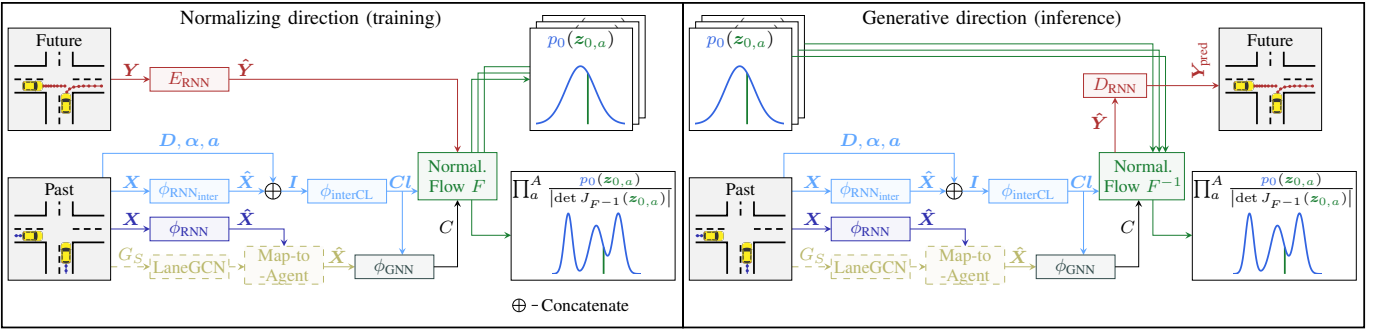}}
    \vspace{-4mm}
    \caption{Architecture of GMoP. 
    During training we use the normalizing direction in which we encode the future trajectories $\bm{Y}$ with $E_{\text{RNN}}$ and transform the abstracted features $\hat{\bm{y}}_a$ to a sample $\bm{z}_{0,a} = F(\hat{\bm{y}}_a)$ assumed to follow a standard normal distribution with the probability density function $p_0$.
    For inference we then use the generative direction, in which a sample $\bm{z}_{0,a} \sim p_0$ is inversely transformed by the Normalizing Flow to generate the abstracted future trajectories $\hat{\bm{y}}_a= F^{-1}(\bm{z}_{0,a})$ that are decoded with $D_{\text{RNN}}$ into the actual trajectories $\bm{Y}_\text{pred}$.
    % The normalizing direction is used during training and the generative direction is used for predicting new trajectories. 
    The likelihood of the encoded scene-level predictions is obtained with $\prod_{a}^{A}p_0 (\bm{z}_{0,a}) \vert \det J_{F^{-1}}(\bm{z}_{0,a})\vert^{-1}$. 
    The encoding $\phi_{\text{CNN}}$ of map $E$, and the scene graph processing modules LaneGCN and Map-to-Agent are optional modules, which can provide richer context information. 
    }
    \vspace{-4mm}
    \label{fig:JointTFarchitecture}
\end{figure*}

\section{GMoP -- \textbf{G}raph-based \textbf{Mo}tion \textbf{P}rediction}
The goal of our approach is to predict a distribution over future agent trajectories $\bm{Y} \in \mathbb{R}^{n_A \times n_O \times 2}$, given past observations of all agents $\bm{X} \in \mathbb{R}^{n_A \times n_I \times 2}$ and static environment information in the form of scene graphs, which together constitute the context $C$. 
Here, $n_A$ denotes the number of agents, and $n_I$ and $n_O$ denote the number of input and output timesteps, respectively.

An important thing to consider when working with normalizing flows is that the number of features need to be preserved between the distribution that one is trying to fit and the base (i.e. latent) distribution of the NF.
To circumvent this issue we leverage the idea of factorizing the joint distribution over all agents in a scene based on a learned interaction graph in the form of a directed acyclic graph (DAG) as introduced in FJMP~\cite{rowe2023fjmp}.
Factorizing the problem in accordance to a DAG enables us to break the problem of learning a complete joint distribution into learning a series of smaller conditional distributions in the form
\begin{equation*}
    p_K(\bm{Y}|C) = {\textstyle \prod_a^A p_K(\bm{y}_a|\bm{Y}_{P,a}, C)},
\end{equation*}
where $\bm{Y}_{P,a}$ are the future trajectories of the parent nodes $P$ of a given agent, $a$. 
As a result, the feature dimension of $\bm{z}_0$ will always be equal to that of $\bm{y}_a$ (where $\bm{y}_a = \bm{z}_K$) for each conditional distribution $p_K(\bm{y}_a|\bm{Y}_{P,a}, C)$, irrespective of the number of agents in the scene.

To obtain this factorization we first need a DAG, based on an interaction graph for the agents in a given scene. 
We take this opportunity to investigate the manner in which the interaction graph can be constructed.

\subsection{Learning the Interaction Graph}
\label{sec:interactionGraphLearning}
To predict the interaction graph we draw inspiration from FJMP~\cite{rowe2023fjmp} and employ a two layer MLP classifier $\phi_\text{interCl}$ with an embedding layer $\phi_\text{em}: \mathbb{R}^f \rightarrow \mathbb{R}^M$ that has an input feature dimensionality of $f$ and embedding dimensionality $M$, and a classification layer $\phi_\text{cl}: \mathbb{R}^M \rightarrow \mathbb{R}^3$ followed by a Softmax activation to ensure the probabilities of the three classes ($m$ influences $n$, $n$ influences $m$, and no interaction) sums to one.

The classifier takes as input the encoded past trajectories of an $m$-$n$ agent-pair $\bm{\hat{x}}_m$, and $\bm{\hat{x}}_n$ - obtained by passing the past trajectories $\bm{x}_m$ and $\bm{x}_n$ through an RNN $\phi_{\text{RNN}_\text{inter}}$ - a one-hot encoding of the agents' types, a distance vector $d_{m,n} = x_{H_m} - x_{H_n}$, and lastly the angle $\alpha_{m,n}$ between the agents' distance vector $d_{m,n}$ and the source agent's (agent $m$ in this case) final displacement $\tilde{x}_m = x_{H_m} - x_{H-1_m}$ obtained as 
\begin{equation*}
    \alpha = \operatorname{acos}\left(\frac{d_{m,n}\cdot\tilde{x}_m}{||d_{m,n}||\cdot||\tilde{x}_m||}\right).
\end{equation*}
The angle $\alpha$ was defined in such a manner to capture interactions between agent pairs in terms of whether agent $m$ is approaching or moving away from agent $n$.
The agent pair information, observed from the perspective of agent $m$, is concatenated into a single vector $i_{m,n} \in \mathbb{R}^f$. 
This is then passed through the classifier $cl_{m,n} = \phi_\text{interCl}(i_{m,n}) = \phi_\text{cl}(\phi_\text{em}(i_{m,n}))$ to obtain the classification $cl_{m,n}$ of the interaction between the $m$-$n$ agent-pair. An overview of the complete architecture used can be found in Fig.~\ref{fig:JointTFarchitecture}.

\begin{figure*}
    \centering
    \resizebox{\textwidth}{!}{\input{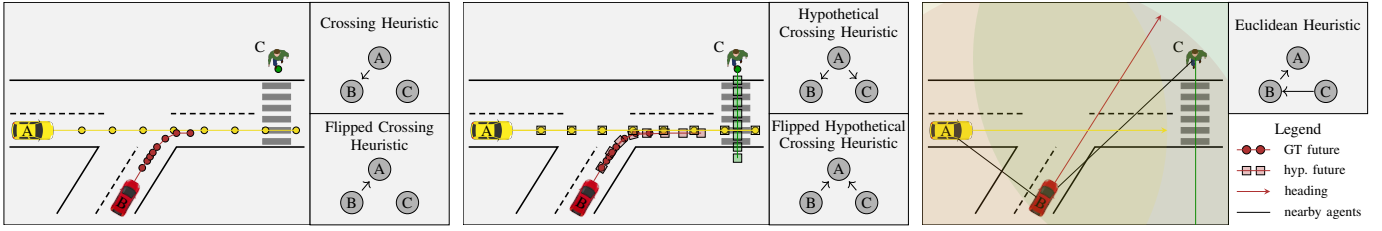}}
    \vspace{-4mm}
    \caption{Illustration of the interaction graphs obtained using the (flipped) crossing, (flipped) hypothetical crossing and Euclidean heuristics. The resulting interaction graphs effectively represent factor graphs from which the joint distribution can be obtained. For example, in the case of the hypothetical crossing heuristic the joint distribution is factorized as $P(A)P(B|A)P(C|A)$ while in its flipped version, it is factorized as $P(B)P(C)P(A|B,C)$. Note, for the hypothetical crossing scenario, the hypothetical future is only used to check whether agents' paths would have crossed had there been no interaction between them. The direction of influence, i.e. interaction, is determined by which agent's GT future reached the hypothetical crossing point first.}
    \vspace{-4mm}
    \label{fig:Heuristics}
\end{figure*}

\subsection{Interaction Graph Heuristics}

\paragraph{\textbf{Euclidean distance}} is one of the most commonly used heuristics for establishing interactions between agents. 
Agents are considered to be interacting if they are within a distance $\epsilon$ of each other in the last past timestep.
Since we need uni-directional edges between agents for factorizing the joint distribution, we expand this heuristic to consider the angle $\phi_{m,n}$ at which an agent $m$ sees an agent $n$.
This angle is obtained as:
\begin{equation*}
    \phi_{m,n} = |\operatorname{atan2}(s_{n,y} - s_{m,y}, s_{n,x} - s_{m,x}) - \gamma_m| \in [0, \pi]
\end{equation*}
where $s_{m, x}$, $s_{n, x}$, $s_{m, y}$, $s_{n, y}$ represent the $x$ and $y$ positions of agents $m$ and $n$ at the last past timestep $H$, and $\gamma_m$ is the heading angle of agent $m$.
We use an angle of $\phi_{m,n} \in [0, \pi]$ since we do not distinguish if an agent is seen on the left or right from the view of the observing agent $m$.
An agent $n$ is influenced by an agent $m$, if the angle $\phi_{n,m}$ at which agent $n$ sees agent $m$ is smaller than $\phi_{m,n}$. 
This then translates to a directed edge going from agent $m$ to $n$.
We do this under the assumption that an agent focuses more on another agent the closer the other agent is to the center of their field of view, i.e. aligned with their heading direction.
We further weigh the interaction using edge weights defined as:
\begin{equation*}
    w_{m,n} = \frac{\epsilon-d(x_{H_m}, x_{H_n})}{\epsilon},
\end{equation*}
where $d(x_{H_m}, x_{H_n})$ is the Euclidean distance between the positions $x_{H_m}$ and $x_{H_n}$ of agents $m$ and $n$ at the last past timestep $H$.
This weighing captures the aspect that the further away an agent $m$ is from agent $n$, the less of an effect this agent has on agent $n$.
It should be noted that for this heuristic, no training is needed, since the construction of the graph follows directly from the above formalization.

\paragraph{The \textbf{crossing heuristic}} used in FJMP~\cite{rowe2023fjmp}, determines the type of interaction between agent pairs from the agents' ground truth future trajectories. 
More concretely, they perform a crossing check between pairs of trajectories.
If agent $m$ reaches the crossing point before agent $n$, then agent $m$ influences agent $n$. 
If there is no crossing point between the trajectories then there is no influence and in turn no edge between the two agents in the interaction graph.
This information was then used to pre-train the interaction graph.

In our approach we formalize this heuristic in the following manner.
Given the trajectory pairs $(\bm{y}_m, \bm{y}_n)$ we calculate a distance matrix $D$ of the form:
\begin{equation*}
    D = \begin{bmatrix}
        d(y_{0_m}, y_{0_n}) & \dots & d(y_{0_m}, y_{T_n}) \\
        \vdots & \ddots & \vdots \\
        d(y_{T_m}, y_{0_n}) & \dots & d(y_{T_m}, y_{T_n})
    \end{bmatrix}
\end{equation*}
for all timesteps of the future trajectories, where $d(y_{t_m}, y_{t_n})$ is the Euclidean distance between the trajectories $\bm{y}_m$ and $\bm{y}_n$ at timestep $t\in[0,T]$.
To establish whether the trajectories cross at any point in time, we check whether any of the distances in $D$ fall under a specified threshold $\epsilon^a_{m}$, dependent on the type of agent $a$ and their average width, since the distances are calculated based on agent center-points. 
More specifically, a crossing is detected when 
$
    d(y_{t_m}, y_{t_n}) \leq \epsilon^a_{m}.
$
We then check the timesteps for the first detected crossing.
If $t_{c,m} < t_{c,n}$, then agent $m$ influences agent $n$ which in the interaction graph is a directed edge going from agent $m$ to agent $n$.
Conversely if $t_{c,m} > t_{c,n}$, the directed edge points from agent $n$ to agent $m$.
If no crossings are detected, there is no interaction between the agents and as such no edge.

A shortcoming of this heuristic is that it assumes agent paths need to cross for an interaction to occur.
However, interactions can also occur in the form of preventing an agent from carrying out their intended behavior.
A simple example of this is a pedestrian standing on the side of a road, intending to cross yet being forced to wait due to vehicles passing by.

\paragraph{\textbf{Hypothetical crossing heuristic}} is a variant of the above heuristic which we define to address the aforementioned shortcoming.
For this we extrapolate a given ground truth future trajectory based on the agent's velocity $\dot{x}_H$ and heading $\theta_H$ at the last timestep of the past trajectory $\bm{x}$.
To ensure realistic trajectories, we use the ground truth future trajectory $\bm{y}$ as a guideline, so as to maintain the overall shape of the trajectory while speeding it up.
The manner in which the speed up is performed is based on two factors.
The first is a check for $\dot{x}^a_H < v^a$, where $v^a$ is the average speed for a given agent type $a$.
The average speeds for the agent types are based on the statistics obtained from the nuScenes~\cite{caesar2020nuscenes} dataset as a guideline for normal speeds in urban environments.
This check captures both agents which are at a standstill as well as agents forced to move at a slower pace than normal.
For these agents, we extrapolate the trajectories such that 
\begin{equation*}
    \forall t\ \dot{y}^a_t < v^a \Rightarrow \dot{y}^a_t = v^a.
\end{equation*}
This ensures, that if at any point in the future the agent's trajectory moves at less than average velocity, it will be sped up to reach the average velocity for its agent type.
In the case that $\dot{x}^a_H > v^a$, the extrapolation is performed such that
\begin{equation*}
    \forall t\ \dot{y}^a_t < \dot{x}^a_H \Rightarrow \dot{y}^a_t = \dot{x}^a_H.
\end{equation*}
This captures the cases where an agent was forced to slow down in the future, potentially due to surrounding agents.
Based on these extrapolated trajectories, a crossing check such as the one performed for the crossing heuristic is applied. 
However, the result of the crossing check is not directly used to define the interactions. 
Instead it is only used as an indication of whether the agents' trajectories might have crossed had the two agents ignored each other.
If so, we then check which agent reached the crossing point first in reality, and use this to define the direction of interaction analogously to the definition of the crossing heuristic.

\paragraph{\textbf{Flipped crossing} and \textbf{flipped hypothetical crossing heuristics}} are variants of the crossing and hypothetical crossing heuristics in which we flip the direction of the interaction edges. 
From a mathematical perspective, both would be valid factorizations of a joint distribution since $P(A,B) = P(A|B)P(B) = P(B|A)P(A)$.
What we aim to study is whether the semantic meaning of the interaction edges can affect the training of the network.

In the case of either the \textbf{(flipped) crossing} or \textbf{(flipped) hypothetical crossing heuristic} the interaction classifier (Sec.~\ref{sec:interactionGraphLearning}) is pretrained using the interaction classes for every agent-pair according to the used heuristic.
A weighted cross-entropy loss is used to account for class imbalance between interacting and non-interacting agent-pairs
\begin{equation*}
    \mathcal{L} = {\textstyle -\sum_c^C w_cy_c \log(p_c)}
\end{equation*}
where $w_c$ is the class weight, $y_c$ the true class probability which is set to 1 for the correct class, and $p_c$ is the predicted class probability.
After training, the interaction classifier and the RNN-encoder $\phi_{\text{RNN}_\text{inter}}$ for the past trajectories are frozen.

\paragraph{\textbf{No heuristic}} is a further case we investigate since there is always the potential that a chosen heuristic is not the correct one for a given task. 
In this case, the neural network has to learn the best graph connections purely from data in accordance to the loss for fitting the final distribution as provided in Sec.~\ref{sec:distFit}.

Additional information in the form of edge weights for the four crossing heuristics and for the case of no heuristics is provided based on the predicted classification probabilities for the three possible cases (no interaction, agent $m$ affects agent $n$, and agent $n$ affects agent $m$). 
The structure of the interaction classifier is described in Sec.~\ref{sec:interactionGraphLearning}.

\paragraph{\textbf{Independence between agents}} is an assumption used as a baseline within our method.
Although we assume there are no interactions in this case, the training loss from Sec.~\ref{sec:distFit} still optimizes for fitting a joint distribution.

\subsection{Fitting the Distribution}
\label{sec:distFit}
It was previously shown that a better fitting over distributions can be achieved by leveraging an auto-encoder to obtain an abstracted representation of the trajectories over which a distribution should be learned~\cite{meszaros2024trajflow}. 
We apply the same strategy of first transforming the trajectories from positions to displacements with 
% \begin{equation}
$\tilde{y}_t = y_t - y_{t-1}$,
% \end{equation}
and then encoding these trajectories using the RNN-encoder $E_\text{RNN}$ of a pretrained RNN-auto-encoder (RNN-AE) to obtain abstracted trajectories $\bm{\hat{y}}$. 
The abstracted trajectories are then passed to the NF. 
The conditional distribution being learned is thus $p_K(\bm{\hat{y}}_a|\bm{\hat{Y}}_{P,a}, C)$.
For our specific implementation, we aggregate the future predictions of the parent nodes of agent $a$ through sum-pooling to obtain $\bm{\hat{Y}}_{P,a}$.

Since we are fitting a joint distribution, we employ a loss over the joint distribution of all agents in a scene:
\begin{equation*}
    \mathcal{L} = -\frac{1}{N}\sum^{N}_{n=1}\sum^{A}_a\log p_0(F(\bm{\hat{y}}_{a,n})) + \log |\det J_{F}(\bm{\hat{y}}_{a,n})|
\end{equation*}
During the final inference, the predicted abstracted trajectories $\bm{\hat{y}}$ are decoded using the RNN-decoder $D_\text{RNN}$ of the RNN-AE to obtain the final predicted trajectory $\bm{y}_{pred}$.

\subsection{Encoding Context Information}
For encoding the context information we use a series of neural networks.
First, we use an RNN encoder $\phi_\text{RNN}$ to encode the past trajectories $\bm{X}$ of all agents in the scene, giving us $\bm{\hat{X}}$.
If a scene-graph $G_S$ is provided by the dataset, this graph is processed by LaneGCN~\cite{liang2020learning} and fused with the encoded past trajectories $\bm{\hat{X}}$ using the Map-to-Agent network from~\cite{liang2020learning}.
The enriched encoding is then further passed through a Graph Neural Network (GNN) $\phi_\text{GNN}$ to encode the interactions between the agents in the scene.
For this GNN we utilize the same graph structure as the one provided by the previously described interaction heuristics.
The output of the GNN is the resulting context vector $C = \phi_\text{GNN}(\bm{\hat{X}})$.
% In the case that environment information $E$ is provided in the form of images, these are processed using a Convolutional Neural Network (CNN) $\phi_\text{CNN}$ and the output is appended to the context vector $C$, resulting in a new context vector $C = \phi_\text{GNN}(\bm{\hat{X}}) \oplus \phi_\text{CNN}(E)$. 
% For the exact implementation of the components introduced in this section we refer the reader to the Appendix.

\section{Experimental Setup}
\subsection{Models}

To study the effect of the interaction graph's structure on the final prediction performance we compare the seven versions of our method~\footnote{Code at: \url{https://github.com/anna-meszaros/GMoP-Experiment-Setup}}:
\begin{itemize}
    \item GMoP assuming independence between agents (indep)
    \item GMoP without heuristics (w/o H)
    \item GMoP with Euclidean distance heuristic (euclH)
    \item GMoP with crossing heuristic (crH)
    \item GMoP with hypothetical crossing heuristic (hcrH)
    \item GMoP with flipped crossing heuristic (flip crH)
    \item GMoP with flipped hypothetical crossing heuristic (flip hcrH).
\end{itemize}

We use three state-of-the-art scene-level trajectory prediction models as baselines:
\begin{itemize}
    \item ADAPT~\cite{aydemir2023adapt} - efficient joint trajectory prediction model.
    \item AutoBots~\cite{girgis2021latent} - transformer based joint prediction model.
    \item FJMP~\cite{rowe2023fjmp} - an architecture which learns a directed acyclic interaction graph to factorize the future joint distribution for a given scene.
\end{itemize}
It is important to note that ADAPT and FJMP only provide a fixed number of deterministic joint predictions, where each prediction is a mode.
Meanwhile, AutoBots outputs a Mixture Model from which predictions can be sampled.

\subsection{Datasets}
We test the methods on four widely used naturalistic datasets with heterogeneous agents, Argoverse~\cite{wilson2023argoverse}, INTERACTION~\cite{zhan2019interaction}, nuScenes~\cite{caesar2020nuscenes}, and rounD~\cite{rounDdataset}.

For testing the models on \textbf{Argoverse}, we use $n_I = 50$ input timesteps and $n_O = 60$ output timesteps with a sampling frequency of \SI{10}{Hz}, resulting in \SI{5}{s} and \SI{6}{s} of past and future data respectively. 
For \textbf{INTERACTION}, we use $n_I = 10$ and $n_O = 30$ with a sampling frequency of \SI{10}{Hz}, resulting in \SI{1}{s} of past and \SI{3}{s} of future data. 
For \textbf{nuScenes}, we use $n_I = 4$ and $n_O = 12$ with a sampling frequency of \SI{2}{Hz}, resulting in \SI{2}{s} of past and \SI{6}{s} of future data. 

For the three datasets above, the datasets were split based on their respectively provided training and validation sets.

Lastly, for the \textbf{rounD} dataset, we use $n_I = 15$ and $n_O = 25$ with a sampling frequency of \SI{5}{Hz}, resulting in \SI{3}{s} and \SI{5}{s} of past and future data respectively. 
For our experiments, we use the scenarios extracted from the original dataset as done in~\cite{schumann2023using}, which focused on the gap acceptance scenario of a vehicle entering the roundabout. 
This in turn results in a more interactive subset of the original dataset.
For this dataset, we employ a leave-one-out training paradigm, in which models would be trained on all but one recording location. 
This results in three unique splits, one for each of the three locations provided in rounD.

To decrease the effect of random parameter initialization on our final analysis of the results, all of the models are trained using 5 different random seeds.

\subsection{Metrics}
\label{sec:JointTFmetrics}
We evaluate the predictions using the following metrics:
\begin{itemize}
    \item \textbf{joint minADE}/\textbf{joint minFDE} - Average/Final $L_2$ distance (in meters) between the best-predicted trajectories for all agents in the scene and the ground truth, based on $6$ predicted samples. 
    The number of samples was chosen based on the fact that both ADAPT and FJMP provide $6$ predictions.
    % Therefore, for the sake of comparability the lower bound of $6$ samples was taken.
    We chose this metric primarily to obtain an interpretable measure of how closely the predictions of the models capture the single ground truth sample available in real-world test cases, since the usefulness of distribution specific metrics such as Negative Log-Likelihood (NLL) is limited when evaluating on singular ground truth samples. 
    Furthermore, since two of the state-of-the-art methods only predict a fixed number of deterministic trajectories instead of distributions this metric also allows us to better compare our methods to the state-of-the-art.
    \item \textbf{joint NLL} - The average NLL
    of the ground truth, based on the joint distribution of predicted trajectories for all agents in the scene.
    This metric provides us with insight on how well the learned distributions fit the ground truth data.
    To obtain the density estimates needed for the NLL metric, we use a non-parametric density estimation approach proposed in~\cite{mészáros2024robust} so as to ensure a more reliable comparison between models. 
    For estimating the predicted trajectory distribution, up to $100$ sampled trajectories were used.
\end{itemize}
For the sake of brevity the above metrics will be referred to as minADE, minFDE, and NLL. 
For performing the experiments we utilize the STEP~\cite{schumann2025step} benchmarking framework.
Note that joint metrics are inherently more challenging than the corresponding single-agent metrics, as they require accurate predictions for all agents in a scene simultaneously.

\section{Results}

Comparing the performance of the different GMoP variants to the state-of-the-art models, we can see that in terms of the distance metrics minADE/minFDE our models fall in the middle of the performance range attained by the three baselines. 
It is important to note that our models are optimized for achieving the best possible distribution fits across the dataset rather than trying to predict a discrete set of  trajectories for a given scenario.
This is also reflected in the NLL metric, where the GMoP variants are able to consistently achieve better results, outperforming even the AutoBots model which also optimizes for distribution fit.

\begin{table}[t!]
    \caption{Argoverse: average results across five random seeds. Per metric, the best performing model is underlined, while bold values highlight the best performing version of GMoP.}
    \vspace{-1mm}
    \centering
    \setlength{\tabcolsep}{5pt}
\begin{tabular}{l|r @{\hskip 1pt} l|r@{\hskip 1pt}l|r@{\hskip 1pt}l}
Models          & \multicolumn{2}{c|}{minADE} & \multicolumn{2}{c|}{ minFDE} & \multicolumn{2}{c}{NLL} \\ \hline

ADAPT        & \underline{$1.89$} &$\scriptstyle{\pm 0.07}$ & \underline{$4.56$} &$\scriptstyle{\pm 0.13}$ & $4.88e^3$ &$\scriptstyle{\pm 357.14}$ \\
AutoBots  & $4.84$ &$\scriptstyle{\pm 0.52}$ & $6.22$ &$\scriptstyle{\pm 0.45}$ & $2.98e^5$ &$\scriptstyle{\pm 7.04e^4}$ \\
FJMP           & $2.38$ &$\scriptstyle{\pm 0.77}$ & $5.96$ &$\scriptstyle{\pm 2.17}$ & $8.08e^4$ &$\scriptstyle{\pm 6.50e^4}$ \\
\hline
 GMoP w/o H & $4.48$ & $\scriptstyle{\pm 4.56}$ & $10.65$ & $\scriptstyle{\pm 10.24}$ & $146.69$ & $\scriptstyle{\pm 324.02}$ \\
 GMoP indep & $2.21$ & $\scriptstyle{\pm 0.05}$ & $5.53$ & $\scriptstyle{\pm 0.11}$ & $-142.95$ & $\scriptstyle{\pm 73.12}$ \\
 GMoP euclH & $5.74$ & $\scriptstyle{\pm 7.06}$ & $13.81$ & $\scriptstyle{\pm 16.64}$ & $493.68$ & $\scriptstyle{\pm 979.89}$ \\
 GMoP crH & $2.21$ & $\scriptstyle{\pm 0.10}$ & $5.48$ & $\scriptstyle{\pm 0.17}$ & $-149.62$ & $\scriptstyle{\pm 213.09}$ \\
 GMoP hcrH & $\bm{2.18}$ & $\scriptstyle{\pm 0.02}$ & $\bm{5.44}$ & $\scriptstyle{\pm 0.07}$ & $178.27$ & $\scriptstyle{\pm 358.92}$ \\
 GMoP flip crH & $2.23$ & $\scriptstyle{\pm 0.02}$ & $5.58$ & $\scriptstyle{\pm 0.05}$ & \underline{$\bm{-161.22}$} & $\scriptstyle{\pm 136.80}$ \\
 GMoP flip hcrH & $2.24$ & $\scriptstyle{\pm 0.05}$ & $5.60$ & $\scriptstyle{\pm 0.11}$ & $236.73$ & $\scriptstyle{\pm 312.31}$ \\
% Ours         & 1.89 &$\scriptstyle{\pm 0.06}$ & 4.54 &$\scriptstyle{\pm 0.15}$ & $-1.68e^2$ &$\scriptstyle{\pm 48.97}$ \\
% Ours w/ crH & 1.86 &$\scriptstyle{\pm 0.03}$ & 4.45 &$\scriptstyle{\pm 0.07}$ & $-4.64e^2$ &$\scriptstyle{\pm 45.10}$ \\
% Ours w/ hcrH & 1.95 &$\scriptstyle{\pm 0.12}$ & 4.65 &$\scriptstyle{\pm 0.25}$ & $-4.57e^2$ &$\scriptstyle{\pm 40.11}$
                             
\end{tabular} 

    \label{tab:JointTF_Argo}
\end{table}

\begin{table}[t!]
    \caption{INTERACTION: average results across five random seeds. Per metric, the best performing model is underlined, while bold values highlight the best performing version of GMoP.}
    \vspace{-1mm}
    \centering
    \setlength{\tabcolsep}{5pt}
\begin{tabular}{l|r @{\hskip 1pt} l|r@{\hskip 1pt}l|r@{\hskip 1pt}l}
Models          & \multicolumn{2}{c|}{minADE} & \multicolumn{2}{c|}{ minFDE} & \multicolumn{2}{c}{NLL} \\ \hline

ADAPT        & $1.21$ &$\scriptstyle{\pm 0.10}$ & $2.52$ &$\scriptstyle{\pm 0.16}$ & $5.27e^3$ &$\scriptstyle{\pm 204.91}$ \\
AutoBots  & \underline{$0.55$} &$\scriptstyle{\pm 0.04}$ & \underline{$1.49$} &$\scriptstyle{\pm 0.05}$ & $1.83e^3$ &$\scriptstyle{\pm 311.50}$ \\
FJMP           & $0.66$ &$\scriptstyle{\pm 0.25}$ & $1.82$ &$\scriptstyle{\pm 0.61}$ & $2.22e^3$ &$\scriptstyle{\pm 1.06e^3}$ \\
\hline
 GMoP w/o H & $0.74$ & $\scriptstyle{\pm 0.12}$ & $2.23$ & $\scriptstyle{\pm 0.35}$ & $-632.79$ & $\scriptstyle{\pm 9.42}$ \\
 GMoP indep & $\bm{0.66}$ & $\scriptstyle{\pm 0.06}$ & $\bm{2.00}$ & $\scriptstyle{\pm 0.16}$ & $-640.51$ & $\scriptstyle{\pm 4.77}$ \\
 GMoP euclH & $0.70$ & $\scriptstyle{\pm 0.10}$ & $2.08$ & $\scriptstyle{\pm 0.27}$ & $-624.13$ & $\scriptstyle{\pm 2.72}$ \\
 GMoP crH & $0.74$ & $\scriptstyle{\pm 0.11}$ & $2.28$ & $\scriptstyle{\pm 0.38}$ & $-636.38$ & $\scriptstyle{\pm 6.05}$ \\
 GMoP hcrH & $0.68$ & $\scriptstyle{\pm 0.05}$ & $2.06$ & $\scriptstyle{\pm 0.16}$ & \underline{$\bm{-640.79}$} & $\scriptstyle{\pm 0.97}$ \\
 GMoP flip crH & $0.78$ & $\scriptstyle{\pm 0.13}$ & $2.39$ & $\scriptstyle{\pm 0.41}$ & $-633.97$ & $\scriptstyle{\pm 5.80}$ \\
 GMoP flip hcrH & $0.70$ & $\scriptstyle{\pm 0.06}$ & $2.09$ & $\scriptstyle{\pm 0.15}$ & $-639.98$ & $\scriptstyle{\pm 1.98}$ \\
% Ours         & 0.59 &$\scriptstyle{\pm 0.06}$ & 1.78 &$\scriptstyle{\pm 0.15}$ & \underline{$-6.38e^2$} &$\scriptstyle{\pm 6.41}$ \\
% Ours w/ crH & 1.45 &$\scriptstyle{\pm 1.69}$ & 3.94 &$\scriptstyle{\pm 4.26}$ & $-5.50e^2$ &$\scriptstyle{\pm 1.65e^2}$ \\
% Ours w/ hcrH & 0.55 &$\scriptstyle{\pm 0.04}$ & 1.62 &$\scriptstyle{\pm 0.10}$ & $-6.26e^2$ &$\scriptstyle{\pm 0.79}$
 % Ours euclH & $0.73$ & $\scriptstyle{\pm 0.08}$ & $2.21$ & $\scriptstyle{\pm 0.22}$ & $-634.21$ & $\scriptstyle{\pm 7.74}$ \\
 
 % Ours hcrH & $0.93$ & $\scriptstyle{\pm 0.21}$ & $2.68$ & $\scriptstyle{\pm 0.57}$ & $-618.97$ & $\scriptstyle{\pm 23.40}$ \\
 % Ours flip hcrH & $1.60$ & $\scriptstyle{\pm 1.89}$ & $4.32$ & $\scriptstyle{\pm 4.62}$ & $-587.23$ & $\scriptstyle{\pm 108.47}$ \\
                             
\end{tabular} 

    \label{tab:JointTF_Interaction}
\end{table}

\begin{table*}[t!]
    \caption{RounD: average results across five random seeds for the three rounD locations. Per metric, the best performing model is underlined, while bold values highlight the best performing version of GMoP.}
    \vspace{-1mm}
    \centering
    \def\arraystretch{1}%  1 is the default, change whatever you need
\setlength{\tabcolsep}{5pt}
\begin{tabular}{l|r @{\hskip 1pt} l|r@{\hskip 1pt}l|r@{\hskip 1pt}l|r @{\hskip 1pt} l|r@{\hskip 1pt}l|r@{\hskip 1pt}l|r @{\hskip 1pt} l|r@{\hskip 1pt}l|r@{\hskip 1pt}l}
                & \multicolumn{6}{c|}{\scriptsize Location 0} & \multicolumn{6}{c|}{\scriptsize Location 1} & \multicolumn{6}{c}{\scriptsize Location 2} \\
\scriptsize Models         & \multicolumn{2}{c|}{\scriptsize minADE} & \multicolumn{2}{c|}{\scriptsize minFDE} & \multicolumn{2}{c|}{\scriptsize NLL} & \multicolumn{2}{c|}{\scriptsize minADE} & \multicolumn{2}{c|}{\scriptsize minFDE} & \multicolumn{2}{c|}{\scriptsize NLL} & \multicolumn{2}{c|}{\scriptsize minADE} & \multicolumn{2}{c|}{\scriptsize minFDE} & \multicolumn{2}{c}{\scriptsize NLL} \\ \hline

 \scriptsize ADAPT    & $\scriptstyle{9.76}$ & $\scriptscriptstyle{\pm 0.38}$ & $\scriptstyle{16.15}$ & $\scriptscriptstyle{\pm 0.60}$ & $\scriptstyle{5.45e^3}$ & $\scriptscriptstyle{\pm 3.10e^3}$ & $\scriptstyle{7.23}$ & $\scriptscriptstyle{\pm 0.46}$ & $\scriptstyle{12.69}$ & $\scriptscriptstyle{\pm 0.95}$ & $\scriptstyle{2.74e^3}$ & $\scriptscriptstyle{\pm 1.02e^3}$ & $\scriptstyle{6.97}$ & $\scriptscriptstyle{\pm 0.27}$ & $\scriptstyle{12.77}$ & $\scriptscriptstyle{\pm 0.86}$ & $\scriptstyle{7.33e^3}$ & $\scriptscriptstyle{\pm 1.00e^4}$ \\
\scriptsize AutoBots  & $\scriptstyle{5.38}$ & $\scriptscriptstyle{\pm 0.50}$ & $\scriptstyle{13.30}$ & $\scriptscriptstyle{\pm 1.24}$ & $\scriptstyle{1.71e^4}$ & $\scriptscriptstyle{\pm 4.76e^3}$ & $\scriptstyle{3.14}$ & $\scriptscriptstyle{\pm 0.01}$ & $\scriptstyle{8.51}$ & $\scriptscriptstyle{\pm 0.09}$ & $\scriptstyle{8.32e^3}$ & $\scriptscriptstyle{\pm 1.08e^3}$ & \underline{$\scriptstyle{2.95}$} & $\scriptscriptstyle{\pm 0.18}$ & \underline{$\scriptstyle{7.73}$} & $\scriptscriptstyle{\pm 0.50}$ & $\scriptstyle{6.54e^3}$ & $\scriptscriptstyle{\pm 1.05e^3}$ \\
 \scriptsize FJMP     & \underline{$\scriptstyle{4.50}$} & $\scriptscriptstyle{\pm 0.49}$ & \underline{$\scriptstyle{11.55}$} & $\scriptscriptstyle{\pm 1.66}$ & $\scriptstyle{6.46e^3}$ & $\scriptscriptstyle{\pm 1.97e^3}$ & \underline{$\scriptstyle{2.74}$} & $\scriptscriptstyle{\pm 0.07}$ & \underline{$\scriptstyle{8.12}$} & $\scriptscriptstyle{\pm 0.24}$ & $\scriptstyle{7.75e^3}$ & $\scriptscriptstyle{\pm 571.68}$ & $\scriptstyle{3.22}$ & $\scriptscriptstyle{\pm 0.37}$ & $\scriptstyle{9.26}$ & $\scriptscriptstyle{\pm 0.71}$ & $\scriptstyle{4.81e^3}$ & $\scriptscriptstyle{\pm 614.46}$ \\
 \hline
 \scriptsize GMoP w/o H & $\scriptstyle{7.23}$ & $\scriptscriptstyle{\pm 2.54}$ & $\scriptstyle{16.36}$ & $\scriptscriptstyle{\pm 4.27}$ & $\scriptstyle{212.01}$ & $\scriptscriptstyle{\pm 399.98}$ & $\scriptstyle{4.30}$ & $\scriptscriptstyle{\pm 1.96}$ & $\scriptstyle{11.49}$ & $\scriptscriptstyle{\pm 5.54}$ & $\scriptstyle{-65.97}$ & $\scriptscriptstyle{\pm 24.96}$ & $\scriptstyle{3.44}$ & $\scriptscriptstyle{\pm 0.34}$ & $\scriptstyle{9.19}$ & $\scriptscriptstyle{\pm 0.98}$ & $\scriptstyle{-53.17}$ & $\scriptscriptstyle{\pm 15.80}$ \\
 \scriptsize GMoP indep & $\scriptstyle{6.33}$ & $\scriptscriptstyle{\pm 0.43}$ & $\scriptstyle{14.89}$ & $\scriptscriptstyle{\pm 0.96}$ & \underline{$\bm{\scriptstyle{-52.83}}$} & $\scriptscriptstyle{\pm 15.31}$ & $\scriptstyle{3.59}$ & $\scriptscriptstyle{\pm 0.23}$ & $\scriptstyle{9.52}$ & $\scriptscriptstyle{\pm 0.51}$ & $\scriptstyle{-70.32}$ & $\scriptscriptstyle{\pm 15.74}$ & $\bm{\scriptstyle{3.37}}$ & $\scriptscriptstyle{\pm 0.13}$ & $\bm{\scriptstyle{8.91}}$ & $\scriptscriptstyle{\pm 0.55}$ & \underline{$\bm{\scriptstyle{-89.47}}$} & $\scriptscriptstyle{\pm 9.36}$ \\
 \scriptsize GMoP euclH  & $\scriptstyle{5.91}$ & $\scriptscriptstyle{\pm 0.23}$ & $\scriptstyle{13.84}$ & $\scriptscriptstyle{\pm 0.55}$ & $\scriptstyle{67.07}$ & $\scriptscriptstyle{\pm 64.59}$ & $\bm{\scriptstyle{3.30}}$ & $\scriptscriptstyle{\pm 0.23}$ & $\bm{\scriptstyle{8.68}}$ & $\scriptscriptstyle{\pm 0.63}$ & $\scriptstyle{-69.60}$ & $\scriptscriptstyle{\pm 29.09}$ & $\scriptstyle{3.41}$ & $\scriptscriptstyle{\pm 0.11}$ & $\bm{\scriptstyle{8.91}}$ & $\scriptscriptstyle{\pm 0.33}$ & $\scriptstyle{-62.41}$ & $\scriptscriptstyle{\pm 21.73}$ \\
 \scriptsize GMoP crH & $\scriptstyle{5.89}$ & $\scriptscriptstyle{\pm 0.46}$ & $\scriptstyle{13.78}$ & $\scriptscriptstyle{\pm 0.97}$ & $\scriptstyle{82.36}$ & $\scriptscriptstyle{\pm 144.12}$ & $\scriptstyle{3.48}$ & $\scriptscriptstyle{\pm 0.15}$ & $\scriptstyle{9.14}$ & $\scriptscriptstyle{\pm 0.46}$ & $\scriptstyle{-61.18}$ & $\scriptscriptstyle{\pm 21.68}$ & $\scriptstyle{3.55}$ & $\scriptscriptstyle{\pm 0.21}$ & $\scriptstyle{9.27}$ & $\scriptscriptstyle{\pm 0.38}$ & $\scriptstyle{-43.32}$ & $\scriptscriptstyle{\pm 29.72}$ \\
 \scriptsize GMoP hcrH & $\bm{\scriptstyle{5.67}}$ & $\scriptscriptstyle{\pm 0.30}$ & $\bm{\scriptstyle{13.44}}$ & $\scriptscriptstyle{\pm 0.80}$ & $\scriptstyle{-3.79}$ & $\scriptscriptstyle{\pm 12.16}$ & $\scriptstyle{3.54}$ & $\scriptscriptstyle{\pm 0.24}$ & $\scriptstyle{9.29}$ & $\scriptscriptstyle{\pm 0.56}$ & $\scriptstyle{-50.08}$ & $\scriptscriptstyle{\pm 23.40}$ & $\scriptstyle{3.56}$ & $\scriptscriptstyle{\pm 0.13}$ & $\scriptstyle{9.59}$ & $\scriptscriptstyle{\pm 0.43}$ & $\scriptstyle{-51.14}$ & $\scriptscriptstyle{\pm 13.38}$ \\
 \scriptsize GMoP flip crH & $\scriptstyle{6.47}$ & $\scriptscriptstyle{\pm 0.47}$ & $\scriptstyle{15.00}$ & $\scriptscriptstyle{\pm 0.83}$ & $\scriptstyle{126.99}$ & $\scriptscriptstyle{\pm 148.71}$ & $\scriptstyle{3.48}$ & $\scriptscriptstyle{\pm 0.29}$ & $\scriptstyle{9.12}$ & $\scriptscriptstyle{\pm 0.72}$ & \underline{$\bm{\scriptstyle{-73.36}}$} & $\scriptscriptstyle{\pm 17.32}$ & $\scriptstyle{3.55}$ & $\scriptscriptstyle{\pm 0.28}$ & $\scriptstyle{9.22}$ & $\scriptscriptstyle{\pm 0.62}$ & $\scriptstyle{-67.26}$ & $\scriptscriptstyle{\pm 17.48}$ \\
 \scriptsize GMoP flip hcrH & $\scriptstyle{6.16}$ & $\scriptscriptstyle{\pm 0.27}$ & $\scriptstyle{14.69}$ & $\scriptscriptstyle{\pm 0.66}$ & $\scriptstyle{-19.43}$ & $\scriptscriptstyle{\pm 27.93}$ & $\scriptstyle{3.60}$ & $\scriptscriptstyle{\pm 0.21}$ & $\scriptstyle{9.50}$ & $\scriptscriptstyle{\pm 0.51}$ & $\scriptstyle{-55.40}$ & $\scriptscriptstyle{\pm 22.48}$ & $\scriptstyle{3.40}$ & $\scriptscriptstyle{\pm 0.11}$ & $\scriptstyle{8.96}$ & $\scriptscriptstyle{\pm 0.30}$ & $\scriptstyle{-65.80}$ & $\scriptscriptstyle{\pm 8.81}$ \\

\end{tabular} 

    \label{tab:JointTF_RounD}
\end{table*}
Looking more closely at versions of GMoP, we can quickly see that out of the different versions there is no one choice that reliably works well for all situations. 
Out of the two large-scale datasets - Argoverse and INTERACTION - we can see that on Argoverse (Tab.~\ref{tab:JointTF_Argo}), all of the heuristics save for the \textbf{Euclidean heuristic} achieve comparable performance in terms of the distance metrics.
We however observe a difference in performance when looking at the distribution fit, in which case the \textbf{flipped crossing heuristic} achieved the best average performance followed closely by the \textbf{crossing heuristic} and the \textbf{independence} assumption.

% Meanwhile, on the INTERACTION dataset we observe that the worst average performance coupled with the highest variability is achieved by the \textbf{(flipped) crossing heuristic} (Tab.~\ref{tab:JointTF_Interaction}).
Meanwhile, on the INTERACTION dataset we observe that the variants of our model perform comparably to each other. 
The worst performance out of the variants, both in terms of average performance as well as higher standard deviation can be seen for the cases with \textbf{no heuristic} as well as the \textbf{(flipped) crossing heuristic}.
At the same time, some of the best performance -- both in terms of average performance as well as low variability across seeds -- is achieved by the \textbf{(flipped) hypothetical crossing heuristic} which accounts for agents not being able to execute their intentions due to the actions of the agents around them.
This is understandable since INTERACTION is a dataset featuring highly interactive and crowded traffic scenarios, meaning that interactions extend beyond the aspect of which agent crosses paths with another agent first.
At the same time, it is noteworthy that similar performance is achieved using the \textbf{independence} assumption.

Looking now at rounD Location 0, we can see that the heuristics (\textbf{Euclidean distance heuristic}, \textbf{(flipped) crossing heuristic}, and \textbf{(flipped) hypothetical crossing heuristic}) all perform reasonably well in terms of distance metrics. 
The \textbf{(flipped) hypothetical crossing heuristic} additionally achieves better NLL compared to the other heuristics.
Meanwhile, the \textbf{independence} assumption, while obtaining worse performance in terms of distance metrics, obtained substantially better performance in terms of NLL. 
Similar observations had been made in~\cite{meszaros2024trajflow}, where better performance in distance metrics did not always imply a better distribution fit and vice versa. 
It is worth noting that the generally poorer performance of both our models and the state-of-the-art baselines on Location 0, compared to the other two locations can be attributed to the difference in dataset sizes generated by the splits, with Location 0 having only 971 training instances, while Locations 1 and 2 were trained with 11220 and 11805 instances respectively.
Nevertheless, the better performance of the \textbf{(flipped) hypothetical crossing heuristic} compared to other versions of our model on this Location, speaks for the fact that adequate heuristics can guide the learning of a model better in the presence of limited training data.
At the same time the better distribution fit achieved when using the \textbf{independence} assumption indicates that attempting to establish structure through heuristics can be highly detrimental to the model if the heuristics fail to capture the interactions between agents.
Meanwhile, on Location 1 we observe comparable performance between variants of our model with the exception of the model with \textbf{no heuristics}.
This model achieved the worst performance in terms of distance metrics both in regards to the average metric values as well as a much higher level of variability.
On Location 2 we see comparable performance between all variants of our model.
The main difference in performance on this location is the performance in regards to the distribution fit, with the best NLL values being obtained through the \textbf{independence} assumption.
% The cause for this could be that even though this subset of the RounD dataset is aimed at the gap acceptance problem it, it still contains a number of scenarios where the degree of interaction is limited such as where both agents are already in the roundabout or the agent in the roundabout is in fact exiting the roundabout.

\begin{table}[t!]
    \caption{NuScenes: average results across five random seeds. Per metric, the best performing model is underlined, while bold values highlight the best performing version of GMoP.}
    \vspace{-1mm}
    \centering
    \setlength{\tabcolsep}{5pt}
\begin{tabular}{l|r @{\hskip 1pt} l|r@{\hskip 1pt}l|r@{\hskip 1pt}l}
Models          & \multicolumn{2}{c|}{minADE} & \multicolumn{2}{c|}{ minFDE} & \multicolumn{2}{c}{NLL} \\ \hline

 ADAPT        & $3.54$ &$\scriptstyle{\pm 0.21}$ & $7.26$ &$\scriptstyle{\pm 0.15}$ & $508.04$ &$\scriptstyle{\pm 115.78}$ \\
AutoBots  & \underline{$3.43$} &$\scriptstyle{\pm 0.07}$ & \underline{$7.06$} &$\scriptstyle{\pm 0.11}$ & $1.01e^3$ &$\scriptstyle{\pm 124.06}$ \\
 FJMP           & $5.68$ &$\scriptstyle{\pm 4.97}$ & $11.53$ &$\scriptstyle{\pm 9.61}$ & $1.56e^5$ &$\scriptstyle{\pm 3.01e^5}$ \\
 \hline
 GMoP w/o H & $5.48$ & $\scriptstyle{\pm 2.16}$ & $12.15$ & $\scriptstyle{\pm 4.29}$ & $446.70$ & $\scriptstyle{\pm 202.09}$ \\
 GMoP indep & $4.43$ & $\scriptstyle{\pm 0.11}$ & $10.19$ & $\scriptstyle{\pm 0.27}$ & $542.31$ & $\scriptstyle{\pm 264.23}$ \\
 GMoP euclH & $5.79$ & $\scriptstyle{\pm 2.92}$ & $13.28$ & $\scriptstyle{\pm 6.70}$ & $371.91$ & $\scriptstyle{\pm 120.77}$ \\
 GMoP crH & $4.31$ & $\scriptstyle{\pm 0.10}$ & $9.80$ & $\scriptstyle{\pm 0.25}$ & \underline{$\bm{348.83}$} & $\scriptstyle{\pm 81.07}$ \\
 GMoP hcrH & $\bm{4.28}$ & $\scriptstyle{\pm 0.09}$ & $\bm{9.74}$ & $\scriptstyle{\pm 0.18}$ & $451.60$ & $\scriptstyle{\pm 111.90}$ \\
 GMoP flip crH & $\bm{4.28}$ & $\scriptstyle{\pm 0.12}$ & $9.77$ & $\scriptstyle{\pm 0.26}$ & $353.31$ & $\scriptstyle{\pm 63.52}$ \\
 GMoP flip hcrH & $4.31$ & $\scriptstyle{\pm 0.10}$ & $9.85$ & $\scriptstyle{\pm 0.20}$ & $510.90$ & $\scriptstyle{\pm 112.76}$ \\
                             
\end{tabular} 

    \label{tab:JointTF_NuScenes}
\end{table}

Meanwhile on nuScenes we once more observe that leveraging heuristics for defining the interactions between agents can help improve performance.  
Specifically, we observe the best performance in terms of distance metrics when using the \textbf{(flipped) crossing heuristic} and \textbf{(flipped) hypothetical crossing heuristics}, and to a lesser extent when using the \textbf{independence} assumption. 
At the same time, the \textbf{(flipped) crossing heuristic} achieve the best distribution fits both in terms of average performance as well as the variability across random seeds, indicating that this heuristic is able to more reliably capture the interactions between agents compared to the other architecture variants.

\section{Conclusion \& Discussion}
Our findings indicate that relying purely on the network to extract interactions between agents based on the training data 
%to then further aid the prediction of their future behaviors 
is often not sufficient for achieving good prediction performance.
In fact, GMoP with no heuristics was generally the poorest performing architecture across our different test cases.
At the same time, using the independence assumption and relying on the joint NLL loss to guide the learning of the network often lead to some of the better performing models.
This is not to say, however, that heuristics are not beneficial for interpreting the interactions between agents.
With the correct heuristics, a clear boost in performance can be achieved as shown on the NuScenes dataset, where the (flipped) crossing heuristic resulted in the best performance between the GMoP variants.
What is further worth noting is that in the case of little training data, choosing a good heuristic which captures the underlying mechanism of interaction in the given scenario can aid in the learning of the model. 
We observed an example of this on RounD Location 0 when using the (flipped) hypothetical crossing heuristics as opposed to the (flipped) crossing heuristics.
What is interesting, is that although the choice of heuristic can have a strong impact on the final performance of the network, the direction of the established edges in the graph does not greatly impact performance.
We can see this in the comparable performance achieved by the (hypothetical) crossing heuristics compared to their flipped counterparts.
This indicates that the presence of the edges is more important than the semantic interpretation of their direction.
However, none of the heuristics provided the best performance across the different datasets, indicating that these heuristics are highly situation dependent. While the (flipped) crossing heuristics generally resulted in better performing models, it was in some cases outperformed by the (flipped) hypothetical crossing heuristics and independence assumption, such as on INTERACTION and RounD Location~0.

\vspace{-0.4mm}
This study acts as a first step towards understanding how to best represent interactions between agents and leverage this for joint predictions.
Nevertheless, there are aspects which our study does not investigate and would be important directions for future research.
For one, we do not consider potential temporal evolution of the interactions themselves.
While the (hypothetical) crossing heuristics strive to capture knowledge about the future evolution of the interaction between two agents it remains a simplification of the continuous negotiations and re-assessments of a situation that traffic participants go through with each other.
Secondly, we make the assumption that the interactions between different agent types are governed by the same mechanisms.
However, this need not be the case particularly when considering interactions between and within vulnerable and non-vulnerable road users.
It would therefore be worthwhile to investigate whether there is a unified mechanism underlying all interactions, or whether interactions should be described differently depending on agents' types.
Thirdly, we only look at the trajectories of the agents to determine their influence on each other and disregard potential confounders such as traffic rules which can have an important impact on the behavior of the agents.
Lastly, while the type of heuristics we used in our study are common to the field of trajectory prediction they are fairly simplified representations of agent interactions. 
Meanwhile, the field of human behavior prediction has been striving towards modeling the underlying mechanisms behind the interaction between agents~\cite{kolekar2020human, siebinga2023modelling}.
These models, while grounded by human behavior studies, are often geared towards specific scenarios such as car following, intersections, merging, etc. and are often also not suitable for real-time multi-modal predictions which are relevant for motion planning.
For this reason, another interesting direction for future research would be to investigate how to combine the benefits of scalability and real-time inference of neural networks with the deeper understanding of human interactions from the field of human behavior modeling.

\label{sec:JointTF_conclusion}

%%%%%%%%%%%%%%%%%%%%%%%%%%%%%%%%%%%%%%%%%%%%%%%%%%%%%%%%%%%%%%%%%%%%%%%%%%%%%%%%

%%%%%%%%%%%%%%%%%%%%%%%%%%%%%%%%%%%%%%%%%%%%%%%%%%%%%%%%%%%%%%%%%%%%%%%%%%%%%%%%

\bibliographystyle{jabbrv_ieeetr}
\bibliography{IEEEabrv,IEEEexample}  % .bib

\end{document}